\documentclass[lettersize,journal]{IEEEtran}
\usepackage{amsmath,amsfonts}
\usepackage{array}
\usepackage[caption=false,font=normalsize,labelfont=sf,textfont=sf]{subfig}
\usepackage{textcomp}
\usepackage{stfloats}
\usepackage{url}
\usepackage{verbatim}
\usepackage{graphicx}
\usepackage{cite}
\usepackage{booktabs} 
\usepackage{times}
\usepackage{epsfig}
\usepackage{graphicx}
\usepackage{amsmath}
\usepackage{amssymb}
\usepackage[dvipsnames, svgnames, x11names]{xcolor}
\usepackage[ruled,vlined]{algorithm2e}
\usepackage{algorithmic}
\usepackage{multirow}
\usepackage{stfloats}
\usepackage{svg}
\usepackage{xcolor}
\usepackage[normalem]{ulem} 
\newcommand\hl{\bgroup\markoverwith
  {\textcolor{yellow}{\rule[-.5ex]{2pt}{2.5ex}}}\ULon}
  
\usepackage{fancyhdr}

\usepackage{multirow}
\usepackage[normalem]{ulem}
\useunder{\uline}{\ul}{}
\usepackage{bbding}
\usepackage{array}

\newcommand{\eg}{\textit{e.g., }}

\begin{document}

\title{Event-based Video Frame Interpolation with Edge Guided Motion Refinement}

\author{Yuhan~Liu,~\IEEEmembership{Graduate Student Member, IEEE,}
        Yongjian~Deng$^{*}$,~\IEEEmembership{}
        Hao~Chen,~\IEEEmembership{} \\
        Bochen~Xie,~\IEEEmembership{} 
        Youfu~Li,~\IEEEmembership{Fellow,~IEEE,}
        and~Zhen~Yang~\IEEEmembership{}
\thanks{Y. Liu, Y. Deng and Z. Yang are with the College of Computer Science, Beijing University of Technology, Beijing 100124, China.} 
\thanks{B. Xie and Y. Li are with Department of Mechanical Engineering,
        City University of Hong Kong, Kowloon, Hong Kong SAR.}
\thanks{H. Chen is with Key Lab of Computer Network and Information Integration (Southeast University), Ministry of Education, Nanjing 211189, China.}  
\thanks{© 20XX IEEE.  Personal use of this material is permitted.  Permission from IEEE must be obtained for all other uses, in any current or future media, including reprinting/republishing this material for advertising or promotional purposes, creating new collective works, for resale or redistribution to servers or lists, or reuse of any copyrighted component of this work in other works.}
\thanks{*: Corresponding author}

}

\markboth{UNDER REVIEW}
{Shell \MakeLowercase{\textit{et al.}}: Bare Demo of IEEEtran.cls for IEEE Journals}

\maketitle
\begin{abstract}
Video frame interpolation, the process of synthesizing intermediate frames between sequential video frames, has made remarkable progress with the use of event cameras. These sensors, with microsecond-level temporal resolution, fill information gaps between frames by providing precise motion cues. However, contemporary Event-Based Video Frame Interpolation (E-VFI) techniques often neglect the fact that event data primarily supply high-confidence features at scene edges during multi-modal feature fusion, thereby diminishing the role of event signals in optical flow (OF) estimation and warping refinement. To address this overlooked aspect, we introduce an end-to-end E-VFI learning method (referred to as EGMR) to efficiently utilize edge features from event signals for motion flow and warping enhancement. Our method incorporates an Edge Guided Attentive (EGA) module, which rectifies estimated video motion through attentive aggregation based on the local correlation of multi-modal features in a coarse-to-fine strategy. Moreover, given that event data can provide accurate visual references at scene edges between consecutive frames, we introduce a learned visibility map derived from event data to adaptively mitigate the occlusion problem in the warping refinement process. Extensive experiments on both synthetic and real datasets show the effectiveness of the proposed approach, demonstrating its potential for higher quality video frame interpolation.
\end{abstract}

\begin{IEEEkeywords}
Event camera, Event-based Video frame interpolation, Neuromorphic vision
\end{IEEEkeywords}
\section{Introduction}
\begin{figure}[t]
\centering
\includegraphics[width=0.9\linewidth]{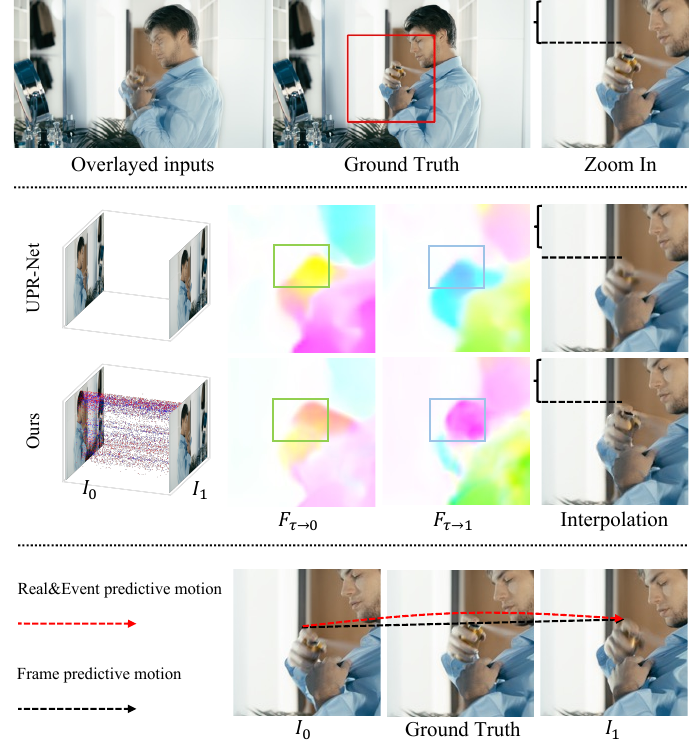}
\caption{Visual comparison between state-of-the-art method UPR-Net\cite{jin2023unified} and ours, where $\{I_0, I_1\}$ are keyframes and \{$F_{\tau \to 0}, F_{\tau \to 1}$\} are predicted bidirectional OFs. Due to the lack of real inter-frame information, traditional VFI work (UPR-Net) cannot accurately model inter-frame motions. Instead, our E-VFI approach can better encode the motion trajectory between consecutive frames.}
\label{fig: first}
\end{figure}

\IEEEPARstart{V}{ideo} Frame Interpolation (VFI) is instrumental in creating high frame-rate videos and providing critical insights into high-speed dynamic situations. It exhibits utility in a wide range of applications, including slow-motion generation \cite{bao2019depth, jiang2018super, xu2019quadratic}, video compression \cite{wu2018video, tip_vc_1}, and video frame prediction \cite{wu2022optimizing, niklaus2017video, chi2020all, choi2020channel, jin2023enhanced, lee2020adacof, tip_vif_1, tip_vif_2}. Yet, traditional VFI approaches struggle to interpolate precisely in complex motion scenarios due to information scarcity between successive video frames, necessitating an assumption of linear motion (Fig. \ref{fig: first}). This dependence heavily compromises the reliability of these studies in practical applications.


\begin{figure}[t]
\centering
\setlength{\abovecaptionskip}{0.cm}
\includegraphics[width=1\linewidth]{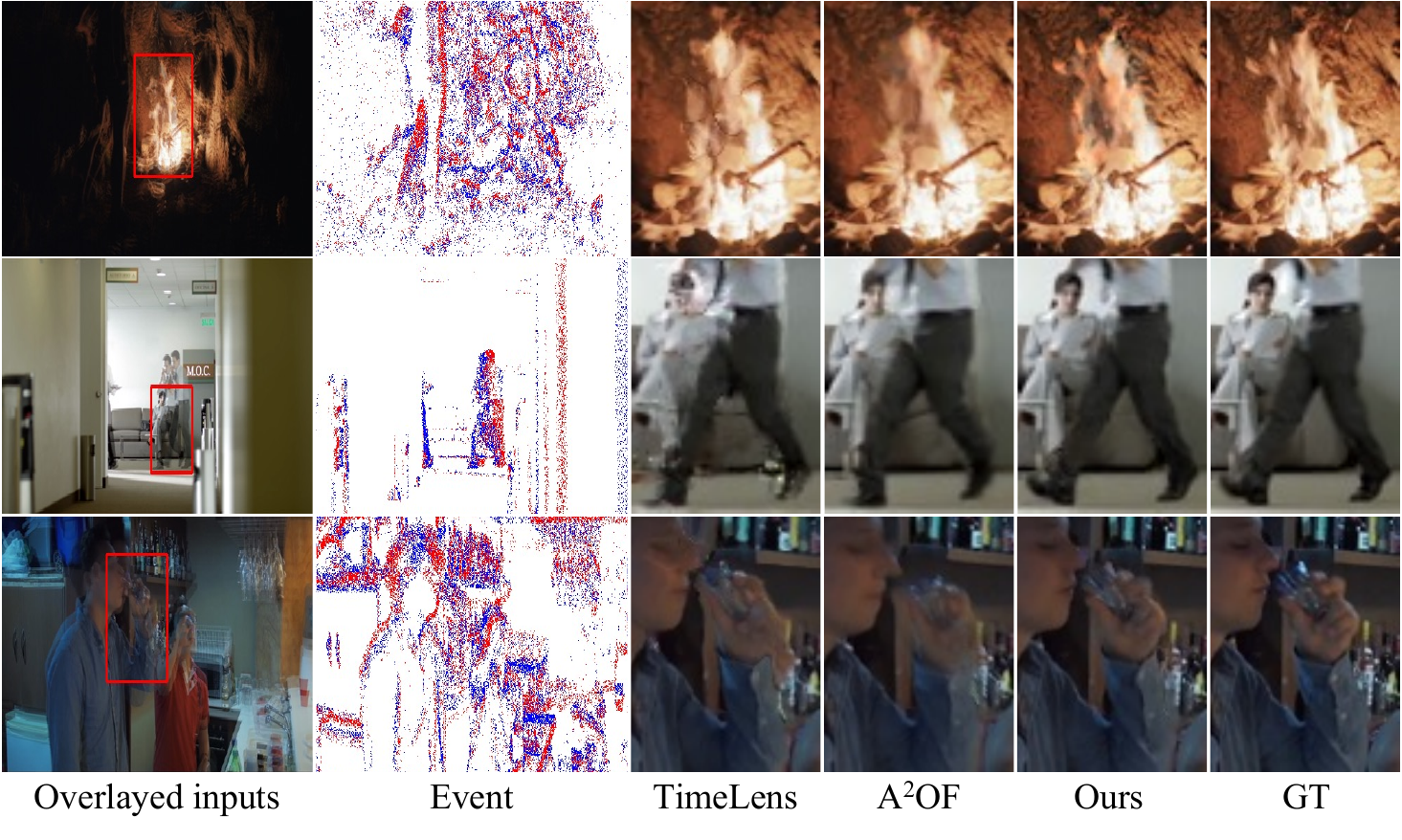}
\caption{Visual comparisons among representative E-VFI methods, the main concerns involve low luminance, irregular motion and textured scenes.}
\label{fig: multi_scene}
\end{figure}

The use of event cameras presents a promising solution to these challenges. By taking advantage of extraordinary temporal resolution (in the order of $\mu s$) \cite{gallego2020event, rebecq2019high}, event data can bridge the information gaps between consecutive frames with precise motion cues (Fig. \ref{fig: first}). Exploiting these benefits, researchers have developed techniques successfully for Event-based Video Frame Interpolation (E-VFI).  Initial research treats events and frames as independent modalities, either employing events solely for motion estimation \cite{tulyakov2021time, he2022timereplayer}, or utilizing events only as a regularization to assist the OF prediction using keyframes \cite{wu2022video}. Despite these studies outperforming traditional VFI works, they diminish the significance of either events or keyframes during interpolation, revealing limitations under extreme scenarios (Fig. \ref{fig: multi_scene}). Recognizing these limitations, certain methodologies \cite{yu2021training, tulyakov2022time, Kim_2023_CVPR} have shifted towards fusing multi-modal features. However, most overlook event signals' inherent properties that mainly provide high-confidence information at scene edges (Fig. \ref{fig: multi_scene}). These works generally handle event representations as traditional visual modalities (\eg RGB and Depth frames), adopting fusion techniques based on existing multi-modal methods \cite{li2016fast, dong2023video}. Such practices may reduce the advantages of events in providing precise motion cues at edges, and amplify the negative impact of noisy events. Therefore, further development in handling event data is necessary to facilitate the potential of E-VFI.

In this study, we aim to address the limitations of prior studies by emphasizing the utilization of high-confidence edge information derived from event data. Specifically, we focus on two core components of the VFI task, namely, motion estimation and warping refinement, by introducing customized modules to achieve this objective. During the motion estimation phase, we present a multi-level attentive fusion mechanism, referred to as the Edge Guided Attentive Module (EGA). The EGA incorporates a Cross-modal Local Attention Module (CLA) designed to identify areas correlated with edge motion, which, with the help of mask aggregation, allows multi-modal OFs to achieve local refinement through neighborhood motion cues. Subsequently, we employ an operation called Cross-OF Attention (COA), which aims to distribute confidence on refined multi-modal features from the CLA, thereby complementing multi-modal motion globally.

During the warping refinement phase, we extend the perspectives from \cite{jiang2018super, huang2022real} to the event modality, introducing an event-based visibility map to address occlusion issues in the VFI task. We first adaptively generate the bidirectional event-based visibility map and combine it with the traditional image-based visibility map for interpolation. Distinct from previous works, we do not generate the visibility map taking advantages of approximated inter-frame motions; instead, we obtain it directly from the original event signals. Given the sensitivity of event cameras to edge information, the event-based visibility map derived from raw events provides depictions of occlusions that occur along moving edges, thereby offering accurate references for keyframe warping and interpolation refinement.

By integrating the above two solutions, we propose a novel E-VFI learning architecture through \underline{E}dge \underline{G}uided \underline{M}otion \underline{R}efinement, namely EGMR. This architecture exploits the accurate event-based motion features at edges to refine the predicted OF used for VFI. Specifically, we estimate motion cues from two modalities using backbone architectures \cite{huang2022real, tulyakov2021time} and optimize multi-modal OFs using the EGA in a coarse-to-fine manner. Finally, we introduce a refinement module that utilizes the event-based visibility map to obtain the final prediction of the interpolated frame.

The contributions of this paper can be summarized as follows: 
\par (1) We reassess the role of event signals in E-VFI tasks, focusing on the feature of event signals that they usually provide reliable information only at the edges of scenes, and introduce a new learning architecture (EGMR) for event-based video frame interpolation. 
\par (2) We propose a cross-modal edge guided attentive module (EGA) for refining the multi-modal OFs, which effectively exploits accurate motion cues at edges provided by events using multi-level attentive mechanisms. 
\par (3) We further incorporate the visibility map generated from sparse event signals into the warping refinement process to address the occlusion challenges in VFI tasks. 
\par (4) Extensive experiments on both synthetic and real E-VFI datasets validate the effectiveness of the proposed learning architecture and our specialized designs that are adapted to the nature of event data.

\section{Related Work}

\subsection{Video Frame Interpolation (VFI)}
The Video Frame Interpolation (VFI) problem has been extensively explored in the literature, with various methods proposed. Four research paths can be distinguished. (1) Motion-based approaches \cite{bao2019depth, bao2019memc, hu2022many, jiang2018super, lu2022video, park2020bmbc, huang2022real, jin2023unified, jin2023enhanced, li2023amt}; (2) Synthesis-based methods \cite{kalluri2023flavr, long2016learning}; (3) Kernel-based \cite{bao2019depth, bao2019memc, cheng2020video} and (4) phase-based \cite{meyer2018phasenet} methodologies. Among these different streams of studies, motion-based approaches that primarily estimate motions between successive frames and generate interpolated frames through warping operations become the most popular choice due to their leading performance and flexible learning structures. For instance, Jiang et al. \cite{jiang2018super} generate multiple intermediate frames using an optical flow (OF) estimation network and a streaming network, and Niklaus et al. \cite{niklaus2018context} improve synthesized frame quality through context-aware modules. Inspired by Park et al. \cite{park2020bmbc}, Jin et al. \cite{jin2023enhanced} designed a recurrent framework that improved complex motion handling. Notably, methods by \cite{jin2023unified} achieve state-of-the-art (SOTA) performance by utilizing  recurrent feature learning and simultaneous update of motions and synthesized images. Yet, while these methods make huge progress in the VFI community, they commonly fail to perform well in extreme situations (\eg complex motion patterns, inconsistent lighting conditions and occlusion) due to blank information between two consecutive frames, resulting in poor reliability and generalization.

\subsection{Event-based Frame Interpolation (E-VFI)}
Early event-based studies prefer to employ uni-modal approaches to accomplish visual tasks, \eg object recognition \cite{Graph-based, 9426390}, semantic segmentation \cite{sun2022ess, 10058930}, video restoration \cite{e2vid, scheerlinck2020fast}, etc. However, the inherent shortcomings of event signals in representing low-contrast regions and scene colors, has rendered them less competent in the VFI task. Thus, researchers commonly accomplish the E-VFI task utilizing complementary of both event data and RGB frames. Benefiting from microsecond-level temporal resolution and HDR characteristics, event cameras can provide accurate motion information between consecutive frames, thereby addressing the challenges posed by the traditional VFI when dealing with lighting inconsistency and non-linear motions. 
Recent successes of E-VFI works \cite{tulyakov2021time, he2022timereplayer, wu2022video, lin2020learning, yu2021training, tulyakov2022time, Kim_2023_CVPR, zhang2022unifying, gao2022superfast} also validate this viewpoint. Initially, works by Tulyakov et al. and He et al. \cite{tulyakov2021time, he2022timereplayer} rely exclusively on events for motion estimation, resulting in inaccurate motion encoding at low texture locations \cite{gao2022superfast}. Confronting this issue, Wu et al. \cite{wu2022video} use the event counting mask to optimize the motion prediction of RGB images, but this diminishes the role of events in motion estimation, manifesting vulnerabilities under extreme conditions. To improve the generalization of learning models, methods proposed in \cite{zhang2022unifying, lin2020learning} tend to train their VFI approaches with an auxiliary task, such as image deblurring. Recent studies \cite{yu2021training, tulyakov2022time, Kim_2023_CVPR} suggest that the key in E-VFI might come from how to effectively exploit two modalities jointly for motion estimation. Specifically, approaches in \cite{yu2021training, tulyakov2022time} introduce new motion estimators that fuse events and frames at multiple scales. Following this path, Kim et al. \cite{Kim_2023_CVPR} enhance the interpolation effect by perform cross-modal complementary on both synthesis and warp processes to further improve the generation effects. However, the sparse and noisy event data typically provide high-confidence motion cues at edges only. Few studies customize their model for such properties of events, which may amplify the negative impact caused by noisy events. Our work acknowledges this issue and utilizes accurate edge motion provided by events as guidance for dense optical flow prediction. We further employ an event-based visibility map in the warping refinement process to address the occlusion problem.

\begin{figure*}[t]
\centering
\includegraphics[width=0.95\textwidth]{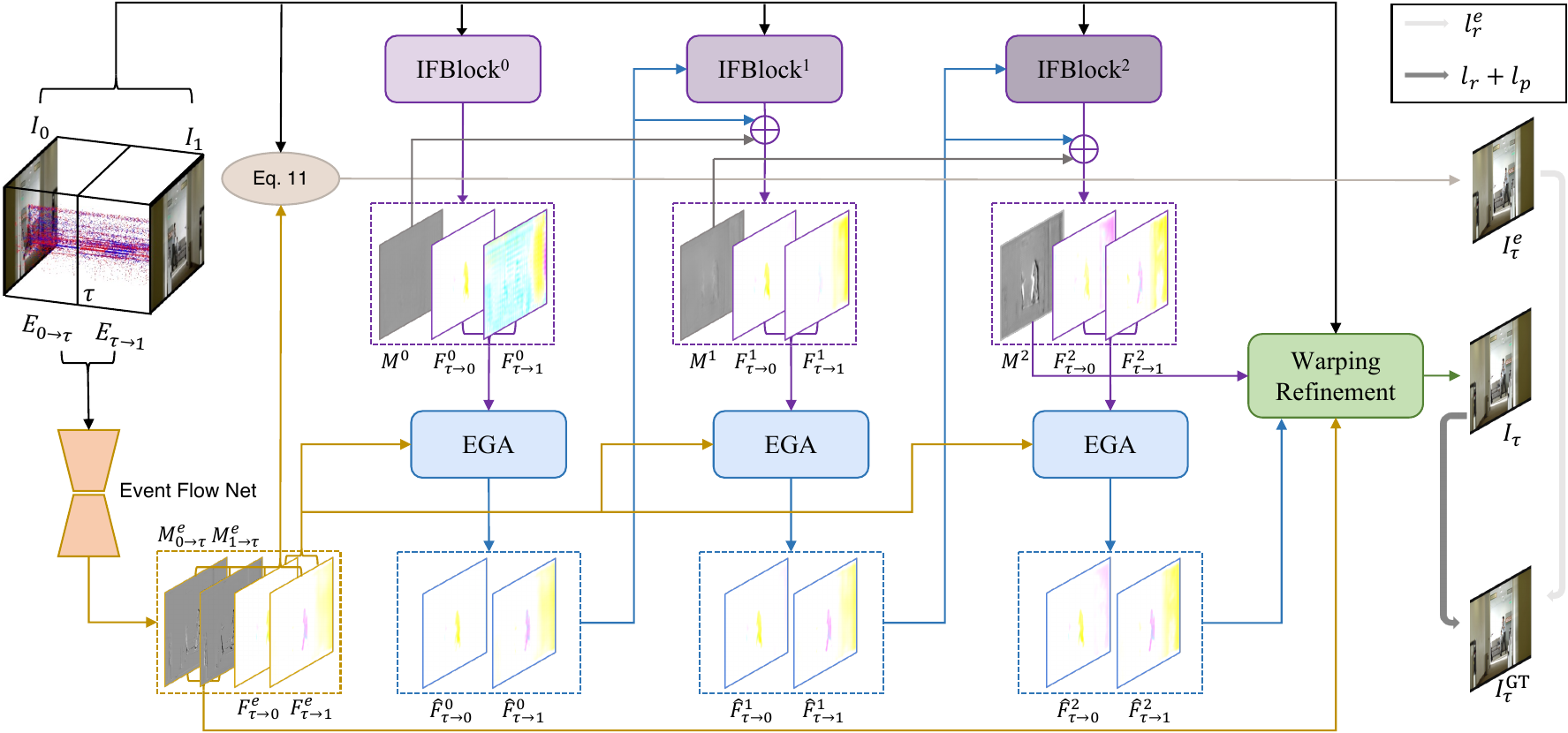}
\caption{Pipeline overview of the proposed EGMR.  Event OF \{$F^e$\} and event-based visibility map \{$M^e$\} are generated from Event Flow Net. Multi-scale frame OF \{$F^{s} | s \in \{ 0,1,2 \}$\} and image-based visibility map ($M^s$) are produced by IFBlocks. At different scales, the EGA is employed to refine predicted multi-modal OFs through emphasizing precise edge motion. Finally, cross-modal enhanced motions are used to generate the final interpolated frame ($I_{\tau}$) via warping and refinement.}
\label{fig: pipline_ifblock}
\end{figure*}
\section{Preliminary and Notations}
\subsubsection{Event Representation}
The $i$-th event $e_i$ in an event stream can be represented as $(x_i, y_i, p_i, t_i)$, where $x_i$ and $y_i$ denote the spatial coordinates, $p_i$ denotes the polarity of the event, and $t_i$ denotes the timestamp of the event. Due to the sparse, noisy, and unstructured nature of the input event stream, the common approach \cite{zhu2019unsupervised, tulyakov2021time, Kim_2023_CVPR} to represent event data is to discretize the time dimension into $B$ consecutive temporal bins and then integrating events into a 3D spatiotemporal Voxel Grid ($E \in \mathbb{R}^{B \times H \times W}$), where  the integration of a specific temporal bin can be formulated as Eq. \eqref{voxelization}.
\begin{equation}
\label{voxelization}
E(k)=\textstyle \sum_{i}^{} p_{i} \max \left(0,1-\left|k-\frac{t_{i}-t_{0}}{t_{N_{e}}-t_{0}}(B-1)\right|\right),
\end{equation}
where $t_{0}$ and $t_{N_{e}}$ respectively denote the start time and end time of the integrated event stream, and $N_{e}$ represents the number of event data. The range of $k$ is in $[0, B-1]$. 
\subsubsection{Problem Formulations}
The objective of our E-VFI approach is to generate interpolated frames ${I}_\tau$ at arbitrary timestamps $\tau \in [0,1]$ using two given reference images $I_0, I_1 \in \mathbb{R}^{3 \times H \times W}$ and inter-frame event data $E_{0 \to 1}$. The input event stream to the network is split into two segments at timestamp $\tau$ and represented as voxel grids ($E_{0 \to \tau}, E_{\tau \to 1} \in \mathbb{R}^{B \times H \times W}$). In this paper, we set the number of time bins $B$ to 5 in our experiments, similar to the setting in \cite{tulyakov2021time}.

\section{Methodology}
\subsection{Overview}
In this work, we propose EGMR for Event-based Video Frame Interpolation (E-VFI) via \underline{E}dge \underline{G}uided \underline{M}otion \underline{R}efinement. The EGMR pipeline is depicted in Fig. \ref{fig: pipline_ifblock}, and is divided into two parts, namely the motion estimation network and the warping refinement module. Specifically, the motion estimation network employs a dual-branch architecture. The first branch houses the Event Flow Net, which approximates sparse event optical flow (${F}^e$) from event representations. The second branch comprises IFBlocks and Edge Guided Attentive modules (EGAs). Here, (a) IFBlocks \cite{huang2022real} are used to calculate dense frame optical flow (${F}$) using reference images and outputs from the previous scale EGA as input; (b) EGAs are designed to improve ${F}$ from IFBlocks by exploiting accurate edge motions from ${F}^e$ and output the optimized optical flow ($\widehat{{F}}$). Finally, the obtained $\widehat{{F}}$, along with visibility maps \{$M, M^e$\} from different modalities, is input into the warping refinement module to produce the interpolated frame ($I_{\tau}$). Subsequently, we detail two crucial components of our learning architecture in sequence, including the motion estimation network (Sec. \ref{MEN}) and the warping refinement module (Sec. \ref{WRM}).

\begin{figure}[t]
\centering
\includegraphics[width=1\linewidth]{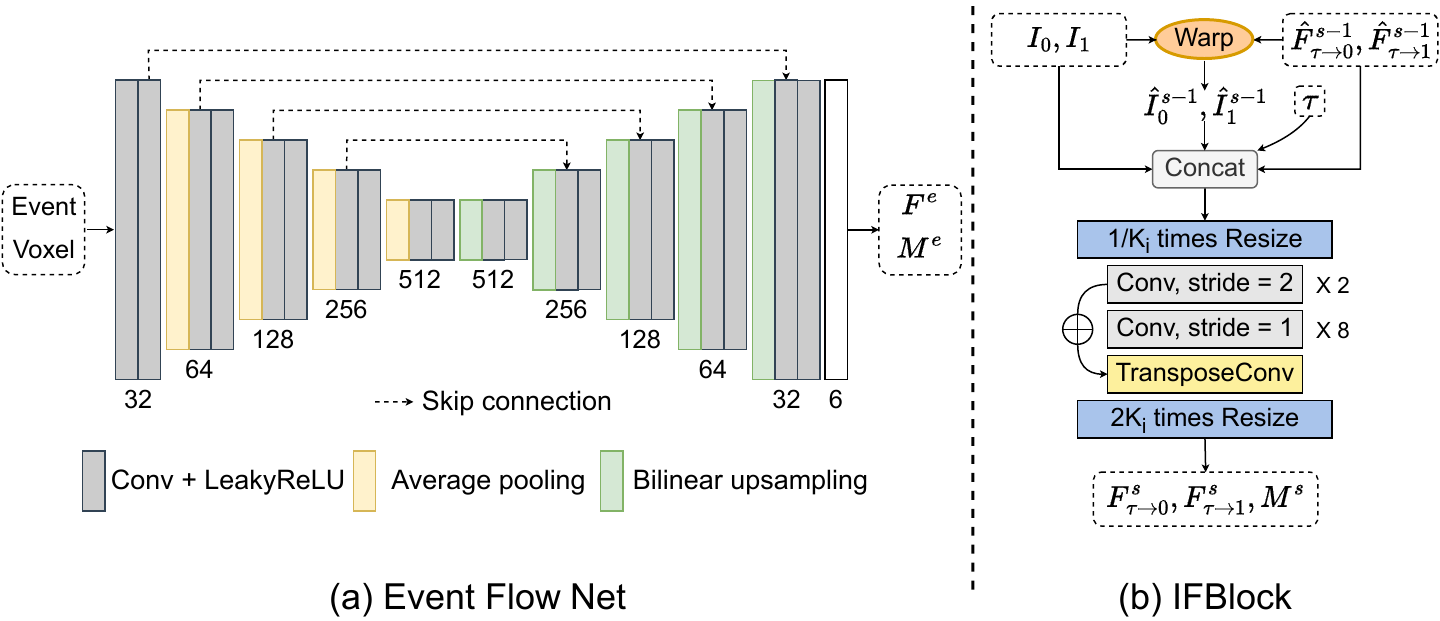}
\caption{The network architecture for extracting event and frame OFs, $K\in[4,2,1]$. The illustration is adapted from \cite{tulyakov2021time, huang2022real}.}
\label{fig: EUnet_IFBlock}
\end{figure}

\subsection{Motion Estimation Network} \label{MEN}
In this section, we outline our motion estimation network, which is divided into three parts. Initially, we give a brief overview of the Event Flow Net and IFBlocks, modules that predict OFs from different modalities. Then, we delve into the core module of our methodology, the Edge Guided Attentive Module, exploring its design motivation and technical details.
\subsubsection{Event Flow Net} 
We utilize a U-Net based network (Fig. \ref{fig: EUnet_IFBlock}.(a)) used in \cite{tulyakov2021time, he2022timereplayer} as our Event Flow Net for generating event OF and event-based visibility map, denoted as  ${F}^e = \left\{F_{\tau\to 0}^{e}, F_{\tau \to 1}^{e}\right\}$ and ${M}^e = \left\{M_{0 \to \tau}^{e}, M_{1 \to \tau}^{e}\right\}$.
By leveraging skip connections between the encoding and decoding components, this module can effectively retains the precise low-level features of event signals in its prediction.
\subsubsection{IFBlock} 
In the context of RGB frames, we incorporate the IFBlock introduced in \cite{huang2022real} to calculate frame OF and image-based visibility map, represented as ${F}^{s} = \left\{F_{\tau \to 0}^{s}, F_{\tau \to 1}^{s} | s \in \left\{0, 1, 2\right\} \right\}$ and $M^s$. We detail the structure of IFBlock in Fig. \ref{fig: EUnet_IFBlock}.(b). This module employs a coarse-to-fine strategy for progressive optimization at different scales $s$. The optimization process is encapsulated in Eq. \eqref{eq:IFBlock}.
\begin{equation}
\label{eq:IFBlock}
F^{s}, M^{s} = IFBlock^{s} (I_0, I_1, \tau, \hat{F}^{s-1})
\end{equation}
where $I_0$ and $I_1$ are keyframes, $\tau$ is the timestamp of the interpolated frame relative to the keyframes, and $\hat{F}^{s-1}$ is the refined OF from the EGA at the previous scale. In general, this progressive procedure can augment OF by observing motion across varying scales. 

\begin{figure}[t]
\centering
\includegraphics[width=0.95\linewidth]{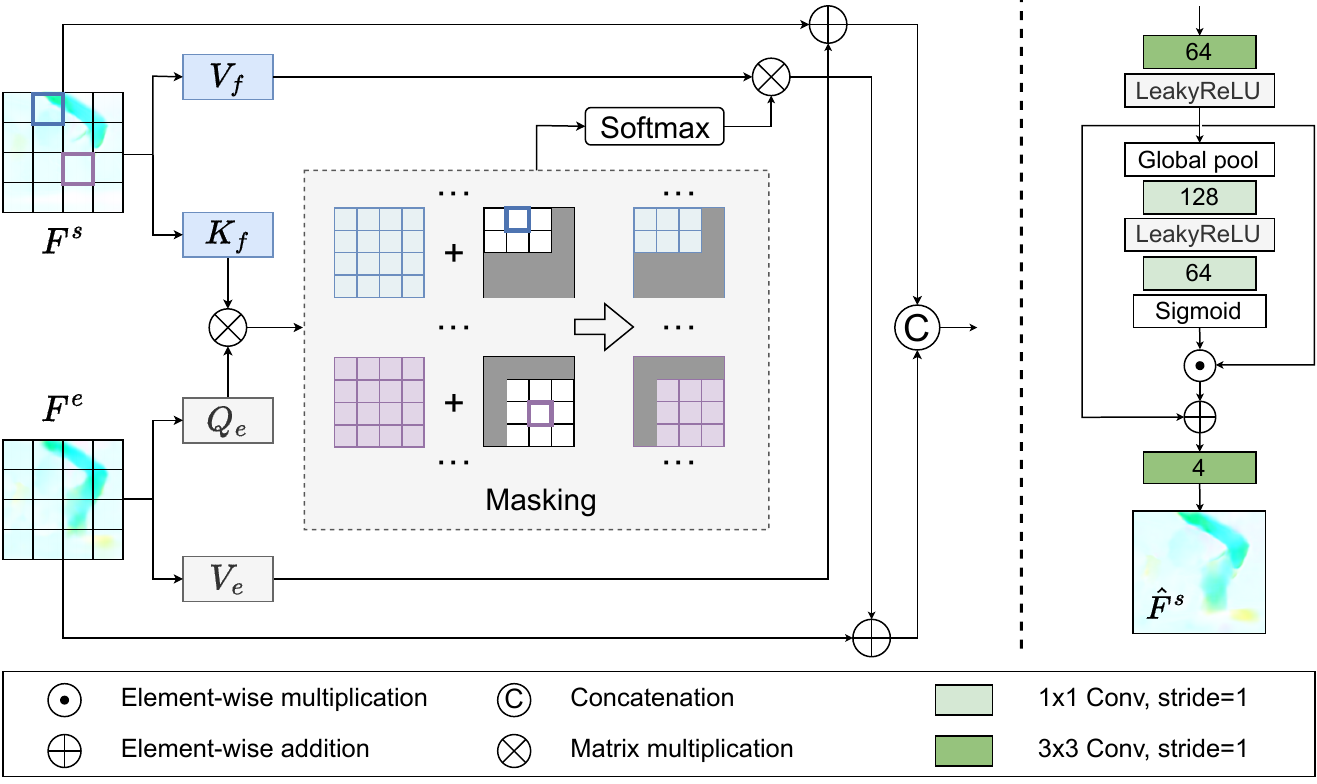}
\caption{Our proposed Edge Guided Attentive module (EGA), left is cross-modal local attention module (CLA) and right is cross-OF attention module (COA). The output of CLA is fed into the module COA. Numbers represent the output channels.}
\label{LFCF}
\end{figure}
\subsubsection{Edge Guided Attentive Module (EGA)}
Previous models often overlook the edge-sensitivity and noisy characteristics of event cameras. This oversight leads to indiscriminate fusion methodologies that might either amplify the noise in the events or underutilize precise, edge-based motion information. To address this issue, we propose the Edge Guided Attentive Module (EGA), a novel approach designed to harness the accurate edge motion derived from the event OF.  As illustrated in Fig. \ref{LFCF}, the EGA employs two key operations: Cross-Modal Local Attention for achieving edge-enhanced optical flow, and Cross-OF Attention for global-level fusion.

\paragraph{Cross-modal Local Attention (CLA)}
We propose the Cross-modal Local Attention (CLA) module to facilitate edge-guided OF augmentation. Through the analysis of imaging and OF prediction tasks across two modalities, we observe that: (1) due to the high temporal resolution and edge sensitivity of event data, event OF accurately encodes edge motion and should remain unadjusted during fusion; (2) event OF is smoother (low-contrast) areas of scenes tends to be impacted by noises largely, making them less unreliable than frame OF; (3) motion cues of a specific location are closely linked with neighboring information.

Upon these observations, we suggest two intuitive design approaches. First, to prevent event OF from being polluted by frame OF, we introduce an asymmetric fusion strategy that optimizes frame OF in a uni-directional manner during cross-modal joint learning. We advocate for local neighborhood connections when optimizing a regional feature in frame OF, instead of performing global fusion as formulated in Eq. \eqref{CLA}.
\begin{equation}
\begin{aligned}
\label{CLA}
V^f_{rect} =Softmax(\frac{Q_eK_f^T+M_l}{\sqrt{d}})V_f,
\end{aligned}
\end{equation}
where $\{Q_e\}$ and $\{K_f, V_f\}$ are respectively obtained by linearly embedding patches from event OF ($F^e$) and frame OF ($F^s$), each of which is sized at $16\times16$. For a particular OF patch at $(x, y)$, we introduce a mask $M_l$, and incorporate it into the global cross-attention map ($Q_e K_f^{T}$), assigning zero to the adjacent region ($\mathcal{R}_a \in \mathbb{R}^{m \times m}$) of the $(x, y)$ location and $-100$ to the distant region.We empirically set $m$ to 3 for the adjacent region $\mathcal{R}_a$ in this study. After applying the \textit{Softmax} operation, the attention scores for the distant region tend to be $0$, thus eliminating their contribution to the feature aggregation process. This approach allows the regions in $F^s$, corresponding to edge motion, to undergo optimization guided by $F^e$, undisturbed by irrelevant motion clues from distant regions. We then add the optimized dense motion feature ($V^f_{rect}$) to the original event OF ($F^e$), thereby supplementing the $F^e$ in low-texture regions while fully preserving event features (Eq. \eqref{FE_sup}).
\begin{equation}
\begin{aligned}
\label{FE_sup}
F^e_{suppl} = F^e + V^f_{rect}.
\end{aligned}
\end{equation}

Considering the noisy nature of event data in low-light, high-speed environments, we realize that the above OF refinement (Eq. \eqref{CLA}) is susceptible to failure in low-texture regions, as it is challenging to judge the event OF is predicted from edges or noisy events. This uncertainty might impairs the motion optimization effects of $V^f_{rect}$. To this end, we incorporate an additional denoising branch in the CLA module, as illustrated in Eq. \eqref{Ff_den}:
\begin{equation}
\begin{aligned}
\label{Ff_den}
F_{smooth}^{s} = V_e + F^{s},
\end{aligned}
\end{equation}
where $\{V_e\}$ is the residual feature embedded from $F^e$, used for enhancing the edge information of $F^s$ to a certain extent. 
Here, we aim to generate another type of edge-enhanced OF ($F_{smooth}^{s}$), which can preserve the original messages of frame OF in low-texture regions as much as possible. The obtained $F_{smooth}^{s}$ is then used for adaptive OF denoising by weighted fusing it with $F^e_{suppl}$, mitigating the adverse impacts introduced by event noise.

\paragraph{Cross-OF Attention (COA)}
Within the CLA process, we prioritize edge-guided dense OF optimization and multi-modal supplementary, resulting in two types of OF, namely $F^e_{suppl}$ and $F_{smooth}^s$. However, the question of how to effectively merge and harness the benefits of these two OF types remains open. As shown in Fig. \ref{LFCF}, we initially employ a convolution operation to fuse these two OF types along the spatial dimensions. Then, drawing inspiration from \cite{yu2023flexible, hu2018squeeze}, we use channel attention (the squeeze-and-excitation layer) to accomplish channel-wise re-weighting operation on fused features from {$F^e_{suppl}$, $F_{smooth}^s$}, considering the global semantics in the scene. With guidance from the Ground Truth, the channel-wise attention can allocate larger weights to discriminative and complementary features for informative co-inference. After local spatial fusion and global channel-wise attention, we use a convolution operation for aggregation, yielding the refined OF ($\widehat{F}^s$), which exploits the advantages of two modalities in both smooth and high-contrast scenarios.

\begin{figure}[t]
\centering
\includegraphics[width=0.9\linewidth]{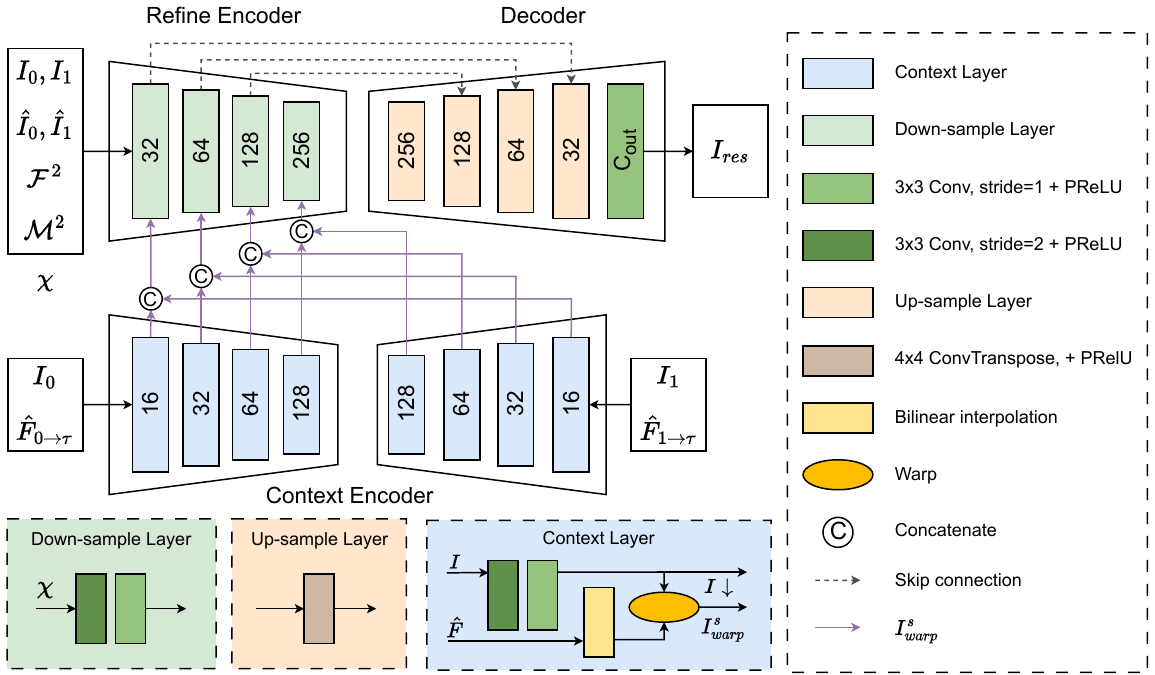}
\caption{The detailed process of refine module. Numbers represent the output channels.}
\label{fig: refine}
\end{figure}
\begin{figure}[t]
\centering
\includegraphics[width=0.9\linewidth]{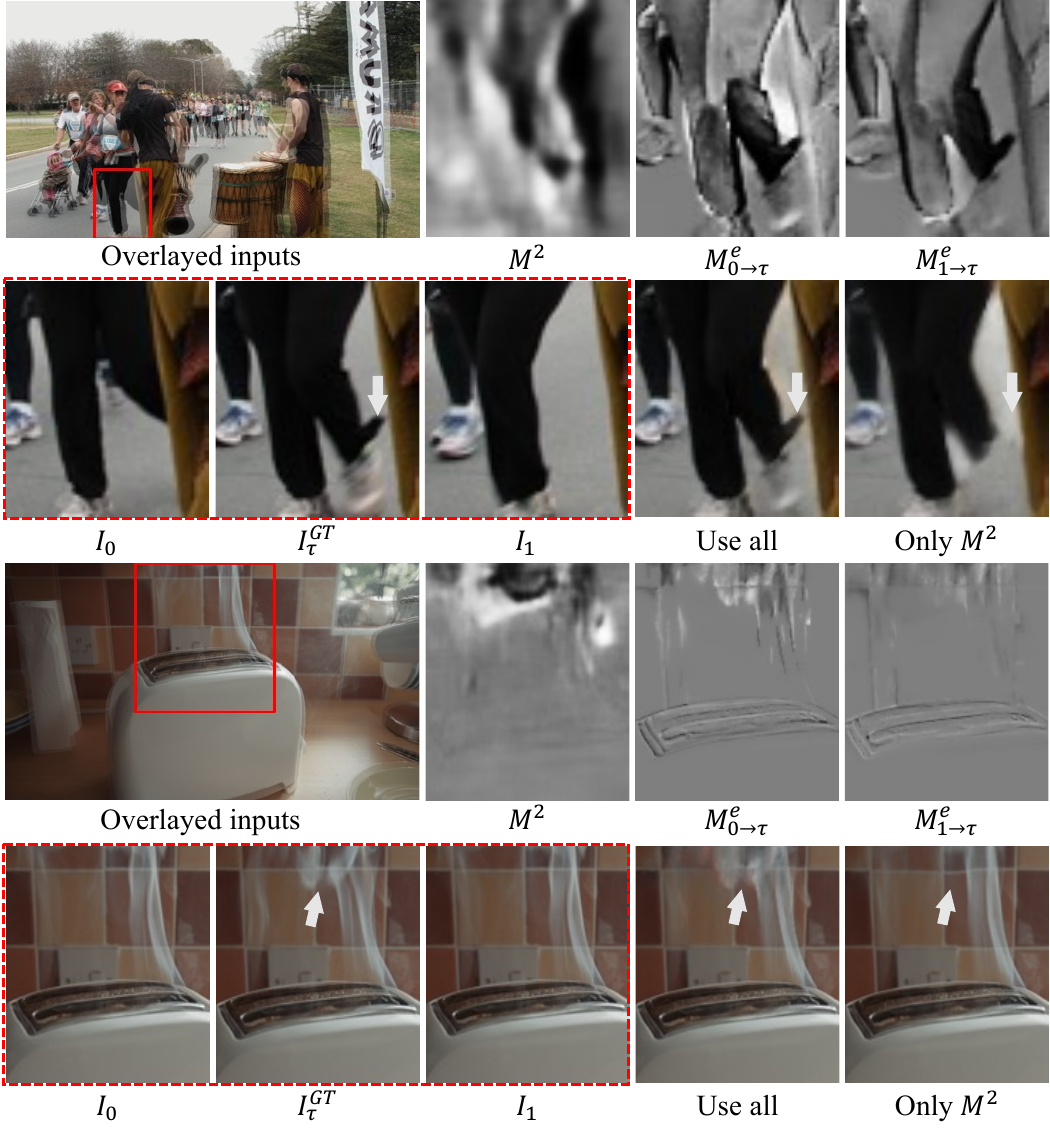}
\caption{Visualization of interpolated frames generated with different visibility maps (\{$M^2, M^e$\}) , where the use of event-based visibility map ($M^e$) in the refine stage enables the reconstruction of moving shoes and bending smoke clearer.}
\label{fig: occlusion_mask_fig}
\end{figure}

\subsection{Warping Refinement Module} \label{WRM}
After obtaining optimized OF ($\widehat{F}^2=\{\hat{F}^2_{\tau\to0}, \hat{F}^2_{\tau\to1}\}$) at the final scale ($s=2$), we apply a backward warping operation to keyframes (\{${I}_{0}, {I}_{1}$\}) to yield two interpolated frames (\{$\hat{I}_{0\to\tau}, \hat{I}_{1\to\tau}$\}). Previous studies \cite{huang2022real, jiang2018super, niklaus2020softmax} have aimed to address the occlusion problem in VFI by generating the image-based visibility map ($M^2$) to fuse interpolated frames (Eq. \eqref{equ: one}). However, due to the inherent low temporal resolution of traditional cameras, the $M^2$ map, which is obtained relying on keyframes, often leads to edge blurring and visual deficiencies in complex motion scenarios (Fig. \ref{fig: occlusion_mask_fig}).

To address these issues, we introduce the event-based visibility map ($M^e = \{M^e_{0\to\tau}, M^e_{1\to\tau}\}$) derived directly from event data to provide accurate inter-frame reconstructing references. As shown in Fig. \ref{fig: occlusion_mask_fig}, the introduced $M^e$ exhibits properties contrary to the traditional $M^2$, showing high sensitivity to motion edge information while underperforming in low-texture areas. These characteristics allow the interpolation influenced by both modalities' visibility maps to be mutually enhancing. Thus, we first use Eq. \eqref{equ: one} and Eq. \eqref{equ: two} to warp the keyframes individually and achieve the fused interpolated frames: $\hat{I}{\tau}^{F}$ and $\hat{I}{\tau}^{E}$.
\begin{equation}
\label{equ: one}
\hat{I}_{\tau}^{F}=M^2 \odot \overrightarrow{\mathcal{W}}(I_{0}, \hat{F}^2_{\tau\to0})+(1-M^2) \odot \overrightarrow{\mathcal{W}}(I_{1}, \hat{F}^2_{\tau\to1}),
\end{equation}
\begin{equation}
\label{equ: two}
\hat{I}_{\tau}^{E}=M^e_{0\to\tau} \odot \overrightarrow{\mathcal{W}}(I_{0}, \hat{F}^2_{\tau\to0})+M^e_{1\to\tau} \odot \overrightarrow{\mathcal{W}}(I_{1}, \hat{F}^2_{\tau\to1}),
\end{equation}
Where $\odot$ denotes element-wise multiplication and $\overrightarrow{\mathcal{W}}$ represents the backward-warping operation. The two preliminary interpolated frames have varying strengths. Namely, $\hat{I}_{\tau}^{F}$  provides smooth results and performs well in low-texture areas, while $\hat{I}_{\tau}^{E}$ preserves precise edge motions. Evidently, the final interpolation frame can be optimized by leveraging the complementary strengths of these two through a fusion operation. To achieve this, we employ a \textbf{convolutional cross-space attention (C$^2$SA)} to merge these two interpolated frames (\{$\hat{I}_{\tau}^{F}, \hat{I}_{\tau}^{E}$\}). As shown in Fig. \ref{fig: refine}, we first expand the dimensions of $\hat{I}_{\tau}^{F}$ and $\hat{I}_{\tau}^{E}$ through convolutions, distinguishing them as queries ($\{Q^F, Q^E\}$) and keys ($\{K^F, K^E\}$) respectively, and then perform cross-attention in the spatial domain. After calculating two spatial similarity matrices, we assign pixel-level fusion weights (${W_0, W_1}$) to the two preliminary interpolated frames \{$\hat{I}_{\tau}^{F}$, $\hat{I}_{\tau}^{E}$\}  using a \textit{Softmax} operation, as formulated in Eq. \eqref{equ: csa}.
\begin{equation}
\label{equ: csa}
\begin{bmatrix}W_0\\ \\W_1\end{bmatrix} = Softmax(\begin{bmatrix}conv(\begin{bmatrix}Q^E,K^F\end{bmatrix})\\ \\conv(\begin{bmatrix}Q^F,K^E\end{bmatrix})\end{bmatrix}),
\end{equation}
\begin{equation}
\label{equ: refine}
I_{\tau}=W_0 \odot \hat{I}_{\tau}^{F} + W_1 \odot \hat{I}_{\tau}^{E} + \Delta I_{\tau},
\end{equation}
where $[\  ]$ denotes concatenation along the channel dimension. Finally, we fuse the interpolated frames $\{\hat{I}_{\tau}^{F}, \hat{I}_{\tau}^{E}\}$ with residual features $\Delta I_{\tau}$ obtained from RefineNet \cite{jiang2018super, niklaus2020softmax, huang2022real}, achieving a final prediction with precise edges and accurately filled smooth visual information (Eq. \eqref{equ: refine}).

\subsection{Losses}
As illustrated in Fig. \ref{fig: pipline_ifblock}, our approach adopt a hybrid loss for model training as formulated in Eq. \eqref{loss}.
\begin{equation}
\label{loss}
\mathcal{L}=\lambda_r^el_r^e + \lambda_rl_r + \lambda_p l_p.
\end{equation}
\noindent\textbf{Reconstruction Loss.} 
$l_r^e$ is used for optimizing Event Flow Net solely, using event OF to warp key frames and calculate reconstruction loss. The event-based interpolated frame $I^e_{\tau}$ is represented as shown in Eq. \eqref{event_warp}, and the computation of $l_r^e$ can be formulated as Eq. \eqref{loss_e}.
\begin{equation}
\label{event_warp}
I_{\tau}^{e}=M^e_{0\to\tau} \odot \overrightarrow{\mathcal{W}}(I_{0}, F^e_{\tau\to0})+M^e_{1\to\tau} \odot \overrightarrow{\mathcal{W}}(I_{1}, F^e_{\tau\to1}),
\end{equation}
\begin{equation}
\label{loss_e}
l^e_r=\|I^{GT}_{\tau}-I_{\tau}^e\|_1.
\end{equation}

As for $l_r$, it is introduced to assess the quality of the intermediate frames' reconstruction, and is calculated according to the following formula:
\begin{equation}
\label{loss_l1}
l_r=\|I^{GT}_{\tau}-I_{\tau}\|_1.
\end{equation}

Both $l_r$ and $l_r^e$ are defined in the RGB space, where pixel values are in the range [0,255].

\noindent\textbf{Perceptual Loss.} We also utilize the perceptual loss \cite{johnson2016perceptual} to enhance semantic consistency, which is defined as:
\begin{equation}
\label{loss_perceptual}
 l_{p}=\frac{1}{N} \sum_{i=1}^{N}\left\|\phi\left(I^{GT}_{\tau}\right)-\phi\left(I_{\tau}\right)\right\|_{2},
\end{equation}
where $\phi$ denotes the $conv4\_3$ features of an pretrained VGG16 model\cite{simonyan2014very}. The proposed model is trained end-to-end by minimizing $\mathcal{L}$, where $\lambda_r, \lambda_r^e, \lambda_p$ are set as $1, 1, 0.1$ respectively. 

\begin{table*}[t]
\centering
\caption{PSNR(dB)/SSIM results on four synthetic datasets. The best results are marked in \textbf{Bold} while the second ones are marked with \underline{underlines}. $\dagger$: we take original video sequences, skip 7 frames, reconstruct them for evaluation. $\ddagger$: retrained with training data identical to ours. \label{tab: sys}}
\renewcommand\arraystretch{1.5}
\setlength{\tabcolsep}{3pt}
\begin{tabular}{lccccccccc}
\toprule[1pt]
\multirow{2}{*}{Method} &
  \multirow{2}{*}{Uses frames} &
  \multirow{2}{*}{Uses events} &
  \multirow{2}{*}{Color} &
  \multirow{2}{*}{Vimeo90k-Triplet} &
  \multirow{2}{*}{Middlebury} &
  \multicolumn{2}{c}{GoPro} &
  \multicolumn{2}{c}{SNU-FILM} \\  \cmidrule(lr){7-8} \cmidrule(lr){9-10}
          &  &  &  &        &        & 7 frames$^\dagger$     & 15 frames      & Hard (7 frames)        & Extreme (15 frames)    \\ \midrule[1pt]
DAIN\cite{bao2019depth}      & \Checkmark    & \XSolidBrush  & \Checkmark  & 34.70/0.944   & 31.33/0.902   & 25.83/0.761 & 24.38/0.758   & 29.39/0.923 & 24.43/0.841 \\
CAIN\cite{choi2020channel}      & \Checkmark    & \XSolidBrush  & \Checkmark  & 34.65/0.948   & 30.94/0.886   & 26.01/0.762 & 24.80/0.746   & 30.05/0.928 & 24.83/0.854 \\
AdaCoF\cite{lee2020adacof}    & \Checkmark    & \XSolidBrush  & \Checkmark  & 34.27/0.945   & 31.17/0.892   & 26.07/0.762 & 24.79/0.746   & 29.19/0.911 & 24.41/0.849 \\
BMBC\cite{park2020bmbc}      & \Checkmark    & \XSolidBrush  & \Checkmark  & 35.06/0.944   & 31.19/0.899   & 25.45/0.755 & 24.29/0.752   & 29.23/0.921 & 23.60/0.833 \\
RIFE\cite{huang2022real}      & \Checkmark    & \XSolidBrush  & \Checkmark  & 34.74/0.957   & 31.54/0.906   & 29.66/0.889 & 25.14/0.772   & 30.36/0.920 & 25.54/0.853 \\
UPR-Net-L\cite{jin2023unified} & \Checkmark    & \XSolidBrush  & \Checkmark  & 36.24/0.966   & 32.68/0.921   & 27.91/0.855 & 24.61/0.758   & 30.82/0.928 & 25.61/0.862 \\ \hline
E2VID\cite{e2vid}     & \XSolidBrush    & \Checkmark  & \XSolidBrush  & 12.70/0.538   & 13.06/0.525   & 12.71/0.384 & 12.45/0.377   & 11.64/0.336 & 11.36/0.356 \\
FireNet\cite{scheerlinck2020fast}   & \XSolidBrush    & \Checkmark  & \XSolidBrush  & 11.36/0.444   & 11.13/0.458   & 10.96/0.271 & 10.90/0.270   & 10.11/0.267 & 10.04/0.264 \\ \hline
TimeLens\cite{tulyakov2021time}  & \Checkmark    & \Checkmark  & \Checkmark  & 36.31/0.962   & \underline{33.27}/0.929   & {34.81/0.959} & {33.21/0.942}   & \underline{31.75/0.935} & \underline{28.64}/0.889 \\
A$^2$OF\cite{wu2022video}$^\ddagger$      & \Checkmark    & \Checkmark  & \Checkmark  & 36.54/0.967   & 32.71/\underline{0.931}   & 34.08/0.954 & 32.65/0.937   & 31.72/0.924 & 28.21/\underline{0.890} \\
CBMNet-L\cite{tulyakov2021time}$^\ddagger$  & \Checkmark    & \Checkmark  & \Checkmark  & \uline{37.69/0.970}   & 31.67/0.897   & \textbf{35.77}/\textbf{0.963} & \textbf{33.95}/\textbf{0.960}   & 29.59/0.885 & 28.08/0.854 \\
\textbf{Ours}      & \Checkmark    & \Checkmark  & \Checkmark  & \textbf{38.35/0.976}   & \textbf{33.54/0.945}   & \uline{35.28}/\textbf{0.963} & \uline{33.51/0.952}   & \textbf{32.76/0.945} & \textbf{30.09/0.912} \\ 
\bottomrule[1.5pt]
\end{tabular}
\end{table*}




\section{Experiments}
\subsection{Experimental Setup}
\subsubsection{Datasets} We train our method on the training set of Vimeo90k-Septuplet \cite{xue2019v90k} following \cite{tulyakov2021time}, where synthetic event data is simulated using the ESIM toolbox \cite{Gehrig2022vid2e}. For evaluation, we follow the evaluation approach adopted in \cite{tulyakov2021time, he2022timereplayer, Kim_2023_CVPR} to test our method on both synthetic and real-world datasets, including Vimeo90k-Triplet \cite{xue2019v90k}, Middlebury \cite{baker2011middlebury}, GoPro \cite{Nah2017gopro}, High Quality Frames (HQF) DAVIS240 \cite{Stoffregen2020reducing}, and High Speed Event and RGB camera (HS-ERGB) \cite{tulyakov2021time}. Furthermore, we include the dataset SNU-FILM \cite{choi2020channel} for performance validation, where we aim to utilize its two most challenging sequences to verify the effectiveness of our approach in complex motion conditions.
  
\subsubsection{Training Settings} Our method is optimized by AdamW \cite{adamw} with weight decay $10^{-4}$ for 100 epochs using PyTorch \cite{paszke2019pytorch}. The initial learning rate is set to $10^{-4}$ and gradually decreased to $10^{-5}$ using cosine annealing. The batch size for each training step was set to 48. During training, we randomly select three frames from a set of seven, with the first and third frames chosen as keyframes ($I_0, I_1$) and the second frame chosen as the ground truth frame ($I^{GT}_{\tau}$) to be interpolated. For data augmentation, we crop the input frames and their paired event voxel grids to a size of $256\times256$ and apply rotation and flipping randomly.

\subsubsection{Metrics} We select PSNR (Peak Signal-to-Noise Ratio) and SSIM (Structural Similarity Index Measure) as our evaluation metrics. (1) PSNR operates on the principle of error sensitivity in image quality assessment, based on the errors among corresponding pixel points. However, PSNR does not account for human visual characteristics, and it often produces evaluation results inconsistent with human perception. (2) SSIM is another full-reference image quality assessment metric that measures image similarity from three aspects: luminance, contrast, and structure. The SSIM index ranges from [0,1], with larger values implying smaller image distortions. In image denoising and similarity evaluation, SSIM often outperforms PSNR, providing results more consistent with human subjective perception.


\begin{figure*}[t]
\centering
\includegraphics[width=0.95\textwidth]{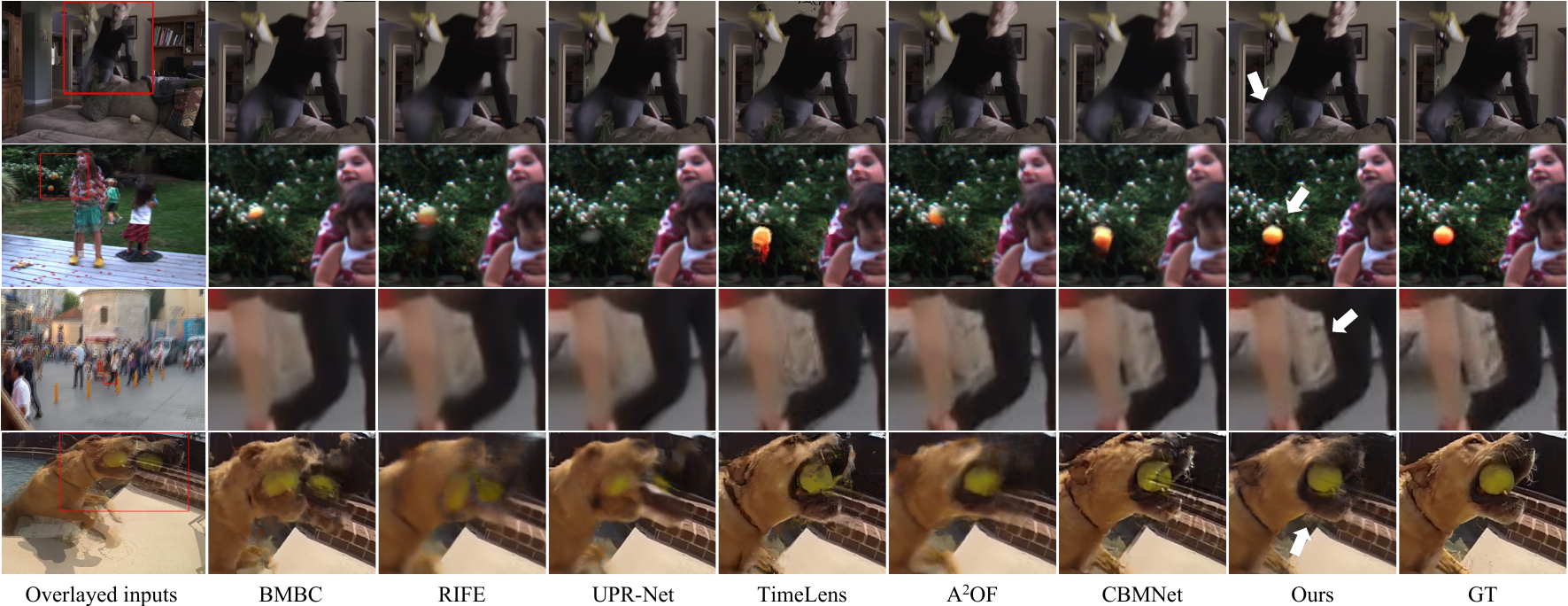}
\caption{Visual comparison among different methods on synthetic datasets. The video clips are from the Vimeo90K-Triplet, MiddleBury, GoPro, and SNU-FILM. We have enlarged the image inside the red box for better display of details.}
\label{fig: 90k_middle_gopro}
\end{figure*}


\begin{table*}[]
\centering
\caption{PSNR(DB)/SSIM results on two real-world datasets, i.e., High Quality Frames (HQF) DAVIS240\cite{Stoffregen2020reducing}, and High Speed Event and RGB camera (HSERGB) \cite{tulyakov2021time}. $\ddagger$: retrained with identical pretraining and finetuning data to ours. \label{tab:real_dataset}}
\renewcommand\arraystretch{1.5}
\setlength\tabcolsep{3pt}
\begin{tabular}{lcccccccccccc}
\toprule[1.5pt]
\multirow{3}{*}{Method}                   & \multirow{3}{*}{Uses frames} & \multirow{3}{*}{Uses events} & \multirow{3}{*}{Color}      &  & \multicolumn{2}{c}{HQF}                                                             &  & \multicolumn{5}{c}{HS-ERGB}                                                                                                                                      \\ \cline{6-7} \cline{9-13} 
                                          &                              &                              &                             &  & \multirow{2}{*}{1 frame}                 & \multirow{2}{*}{3 frame}                 &  & \multicolumn{2}{c}{close}                                                     &  & \multicolumn{2}{c}{far}                                                       \\ \cline{9-10} \cline{12-13} 
                                          &                              &                              &                             &  &                                          &                                          &  & 5 frames                               & 7 frames                               &  & 5 frames                               & 7 frames                               \\ \midrule[1pt]
DAIN\cite{bao2019depth}                        & \Checkmark    & \XSolidBrush  & \Checkmark  & & 30.27/0.879                              & 26.92/0.722                              &  & 29.03/0.807                           & 28.50/0.801                           &  & 27.92/0.780                           & 26.13/0.747                           \\
CAIN\cite{choi2020channel }                      & \Checkmark    & \XSolidBrush  & \Checkmark  & & 30.47/0.871                              & 26.44/0.714                              &  & 29.22/0.821                           & 28.37/0.798                           &  & 26.68/0.767                           & 25.57/0.731                           \\
AdaCoF\cite{lee2020adacof}                       & \Checkmark    & \XSolidBrush  & \Checkmark  & & 30.02/0.859                              & 27.04/0.736                              &  & 29.16/0.808                           & 28.72/0.813                           &  & 27.12/0.772                           & 25.90/0.732                          \\
BMBC\cite{park2020bmbc}                          & \Checkmark    & \XSolidBrush  & \Checkmark &  & 30.72/0.881                              & 27.09/0.741                              &  & 29.22/0.820                           & 27.98/0.807                           &  & 25.62/0.741                           & 24.14/0.710                           \\
RIFE\cite{huang2022real}                         & \Checkmark    & \XSolidBrush  & \Checkmark &  & 31.70/0.889                              & 27.93/0.796                              &  & 33.12/0.857                           & 32.32/0.846                           &  & 29.47/0.849                           & 27.20/0.801                           \\
UPR-Net-L\cite{jin2023unified }                  & \Checkmark    & \XSolidBrush  & \Checkmark &  & 32.15/0.915                              & 27.96/0.863                              &  & 32.22/0.841                           & 31.01/0.829                           &  & 28.85/0.841                           & 26.27/0.787                           \\ \hline
E2VID\cite{e2vid}                                & \XSolidBrush  & \Checkmark    & \XSolidBrush & & 11.91/0.442                              & 12.02/0.445                              &  & 8.87/0.396                            & 9.03/0.399                            &  & 11.46/0.508                           & 11.43/0.505                           \\
FireNet\cite{scheerlinck2020fast}                & \XSolidBrush  & \Checkmark    & \XSolidBrush & & 11.98/0.453                              & 12.08/0.443                              &  & 8.92/0.398                            & 9.06/0.400                            &  & 11.51/0.509                           & 11.49/0.505                           \\ \hline
TimeReplayer\cite{he2022timereplayer} & \Checkmark    & \Checkmark    & \Checkmark  & & 31.07/0.931                              & 28.82/0.866                              &  & 31.21/0.818                           & 29.83/0.816                           &  & 31.98/0.861                           & 30.07/0.834                           \\
CBMNet-L\cite{Kim_2023_CVPR}   & \Checkmark    & \Checkmark    & \Checkmark &  & \underline{34.77}/0.953 & {33.08/0.940} &  & -                                     & -                                     &  & -                                     & -                                     \\
TimeLens\cite{tulyakov2021time}                  & \Checkmark    & \Checkmark    & \Checkmark  & & 33.42/0.934             & 32.27/0.917                              &  &32.19/0.839 &31.68/0.835                           &  &\uline{33.13}/0.877 &\uline{32.31}/0.869                           \\
A$^2$OF\cite{wu2022video}                        & \Checkmark    & \Checkmark    & \Checkmark &  & 33.94/0.945 & 31.85/0.932                              &  & 33.21/0.865 &32.55/0.852                           &  &\textbf{33.64}/\uline{0.891} &\textbf{33.15}/\uline{0.883}                           \\\hline

TimeLens\cite{tulyakov2021time}$^\ddagger$                  & \Checkmark    & \Checkmark    & \Checkmark  & & 33.45/0.933                              & 32.14/0.920                              &  & 32.34/0.845                           & 31.42/0.841                           &  & 32.24/0.869                           & 31.21/0.858                           \\
A$^2$OF\cite{wu2022video}$^\ddagger$                        & \Checkmark    & \Checkmark    & \Checkmark &  & 33.36/0.931                              & 31.27/0.922                              &  & 32.49/0.855                           & 31.23/0.846                          &  & 32.38/0.877                           & 31.42/0.864                           \\
CBMNet-L\cite{Kim_2023_CVPR}$^\ddagger$   & \Checkmark    & \Checkmark    & \Checkmark &  & 34.71/\underline{0.957} & \underline{33.29}\underline{/0.948} &  & \underline{34.47}/\underline{0.863}                                     &  \uline{34.06/0.865}                                    &  & 31.43/0.888                                     & 30.47/0.870                                     \\
\textbf{Ours}            & \Checkmark    & \Checkmark    & \Checkmark  & & \textbf{35.14/0.958}    & \textbf{33.76/0.951}    &  & \textbf{34.88}/\textbf{0.871} & \textbf{34.45/0.869} &  & {32.54}/\textbf{0.934} & {31.79}/\textbf{0.927} \\ \bottomrule[1.5pt]
\end{tabular}
\label{hqfunfinetuning}
\end{table*}

\begin{figure*}[t]
\centering
\includegraphics[width=0.95\textwidth]{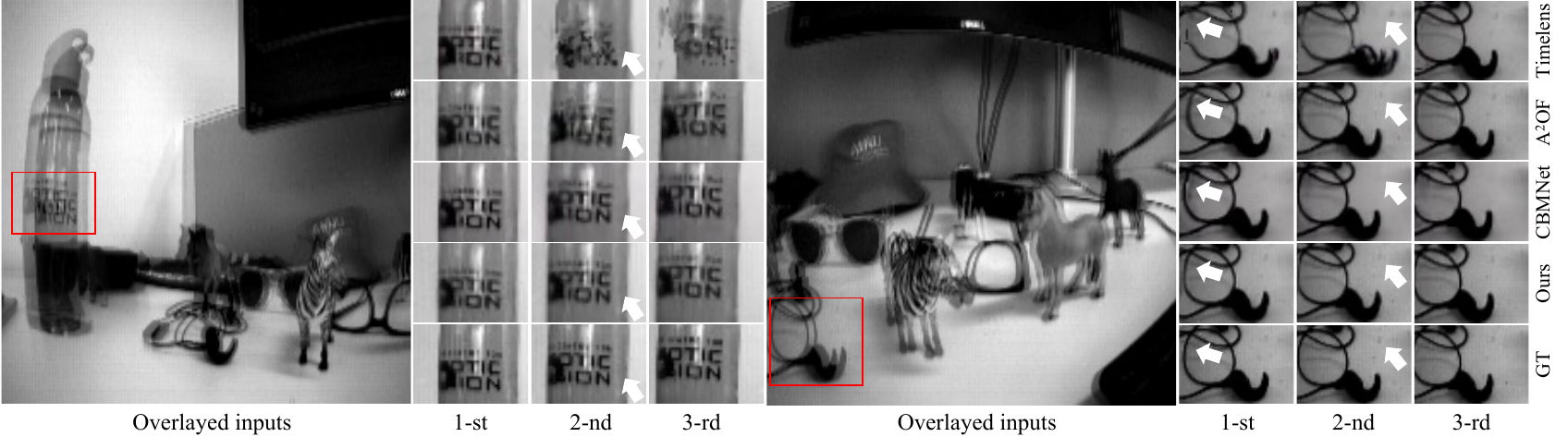}
\caption{Visual comparison on HQF (3 frames skipped). From top to bottom: TimeLens, A$^2$OF, CBMNet, EGMR (ours) and GT.}
\label{hqf_3}
\end{figure*}

\subsection{Comparison to the State-of-The-Art Methods}
We compare our method with state-of-the-art techniques across three categories. (1) Conventional VFI techniques encompassing DAIN \cite{bao2019depth}, CAIN \cite{choi2020channel}, AdaCoF \cite{lee2020adacof}, BMBC \cite{park2020bmbc}, RIFE\cite{huang2022real} and UPR-Net\cite{jin2023unified}. (2) Event-based video reconstruction methodologies including E2VID \cite{e2vid} and FireNet \cite{scheerlinck2020fast}. (3) Event-based VFI strategies including TimeLens \cite{tulyakov2021time}, A$^2$OF \cite{wu2022video}, TimeReplayer \cite{he2022timereplayer}, and CBMNet \cite{Kim_2023_CVPR}. It's worth noting that these models are trained with different datasets in their original paper compared to ours. For instance, A$^2$OF, TimeReplayer and CBMNet use the collection of GoPro, Adobe240 as training data where hold advantages in data abundance to ours. To ensure a fair comparison, we re-train the methods where code is available such as A$^2$OF, TimeLens, and CBMNet, employing optimization settings identical to those in their paper, but with our training and fine-tuning data. For TimeReplayer, we directly cite their performance on real datasets (HQF and HS-ERBG) from their papers as references.

\subsubsection{Evaluations on Synthetic Datasets} 
\paragraph{Quantitative Evaluations} We first evaluate the performance on four synthetic datasets, as depicted in Tab. \ref{tab: sys}. Here, we re-train A$^2$OF and CBMNet using the same training datasets (Vimeo90k-Septuplet) as ours to have a fair comparison. For TimeLens, as this method are trained with identical data to ours, we directly cite their reported results. Tab. \ref{tab: sys} reveals that our approach achieves the highest VFI performance on most compared datasets. First, we evaluate our model on the Vimeo90k-Triplet, Middlebury, and GoPro datasets, where our method outperforms or achieves comparative results in terms of PSNR and SSIM compared to these approaches, highlighting its robustness and versatility across diverse content types under normal motion conditions. 

Then, we include the hard and extreme scenarios in SNU-FILM for evaluation. These scenarios are considered challenging for the VFI task due to all samples contain large and complex motions with long duration. It can be observed that our method achieves the best performance over other works. Notably, our method significantly improves the interpolation performance in extreme scenarios, achieving improvements of 1.45dB and 0.023 in PSNR and SSIM over the second-best method, indicating our method's adaptability to samples with complex motions.

\paragraph{Qualitative Evaluations} 
Fig. \ref{fig: 90k_middle_gopro} displays the visual results on four compared synthetic datasets. Several challenging scenes are selected to illustrate the visual comparison across various approaches. From Fig. \ref{fig: 90k_middle_gopro}, we can see that our proposed method can preserve structural information of the inter-frame accurately, such as moving balls, shaking legs and dog heads, validating the reasonably of our edge-enhanced philosophy in the model's design.


\subsubsection{Evaluations on Real Datasets} 
\paragraph{Quantitative Evaluations}
In this section, we aim to assess our method using the real datasets HQF and HS-ERGB, where the sample distributions in these datasets are different from those in the synthetic datasets. We evaluate our method after fine-tuning on real event data following \cite{tulyakov2021time, wu2022video, he2022timereplayer, Kim_2023_CVPR}. However, after carefully investigation about these studies, there is no detail description about how to fine-tune their models no matter which data are exploited or the scale of data is utilized. Hence, to have a fair comparison, we include models' performance from two perspectives. (1) We retrain code available methods (TimeLens, A$^2$OF \& CBMNet) using the same pre-training and fine-tuning settings. In particular, we randomly select $10\%$ of samples in HQF and HS-ERGB for fine-tuning and the rest for evaluation. (2) We directly cite the reported results for all compared E-VFI works. 

First, we compare our approach with the retrained models. As shown in Tab. \ref{tab:real_dataset}, it is evident that our model consistently outperforms across all conditions and datasets, indicating better image quality preservation and superior structural information reconstruction capabilities compared to TimeLens, A$^2$OF, and CBMNet under the same training strategy.

Next, we measure the performance of our model against the results cited in previous works. As indicated in Tab. \ref{tab:real_dataset}, our method achieves the best performance on HQF and comparable results on the HS-ERGB. In particular, while achieving leading results on the Close-away subset of HS-ERGB with respect to PSNR/SSIM, the PSNR metric of our model on the Far-away subset is lower than the performance reported in the TimeLens and A$^2$OF papers. Given that our method shows advantages compared to these two studies when using the same training settings, we attribute this phenomenon to the different pre-training and fine-tuning data usage. Interestingly, our method shows significant improvements in SSIM on HS-ERGB (far), with a minimum improvement of 0.043 compared to the second-best result, suggesting our method's superiority in preserving structural similarity during interpolation.

\begin{figure*}[t]
\centering
\includegraphics[width=0.95\textwidth]{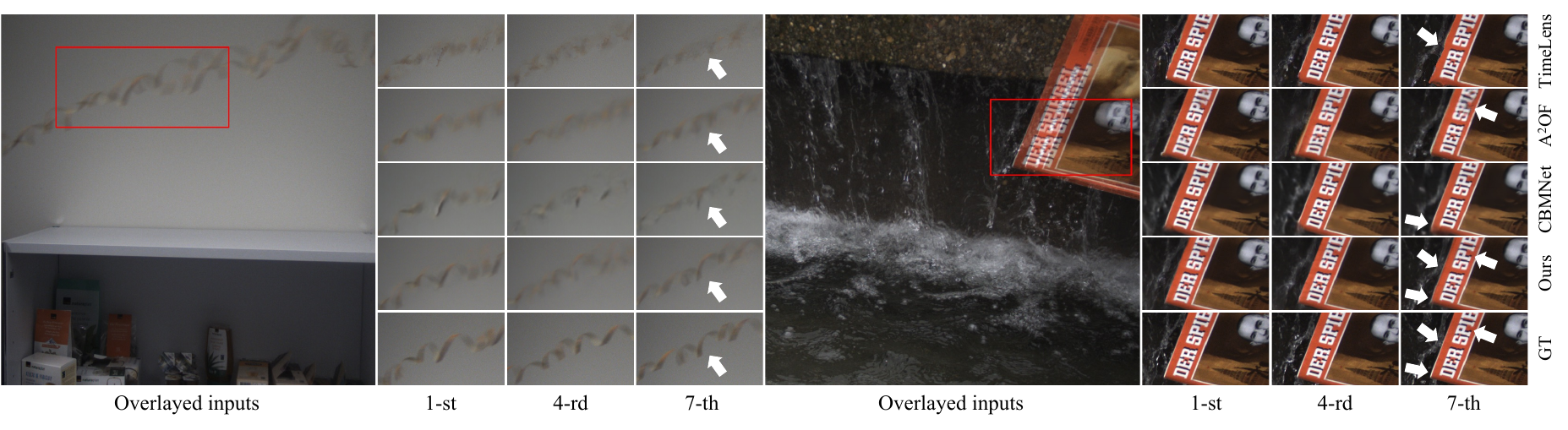}
\caption{Visual comparison on HS-ERGB (7 frames skipped). From top to bottom: TimeLens, A$^2$OF, CBMNet, EGMR (ours) and GT.}
\label{hsergb_5}
\end{figure*}

\paragraph{Qualitative Evaluations} 

Fig. \ref{hqf_3} showcases our approach's superior visual results on the real event dataset HQF, offering clearer and more distinct imagery compared to the distorted output of TimeLens, A$^2$OF and CBMNet. Similarly, Fig. \ref{hsergb_5} demonstrates the efficacy of our method on HS-ERGB, successfully reconstructing sharper motion in fast and irregular scenarios, unlike TimeLens and A$^2$OF which produce blurred and artifact-ridden results (\eg TimeLens  \& CBMNet distorts the book's edges, and A$^2$OF blurs the text).

\subsection{Stress Test}
\begin{figure*}[!t]
\centering
\includegraphics[width=0.95\textwidth]{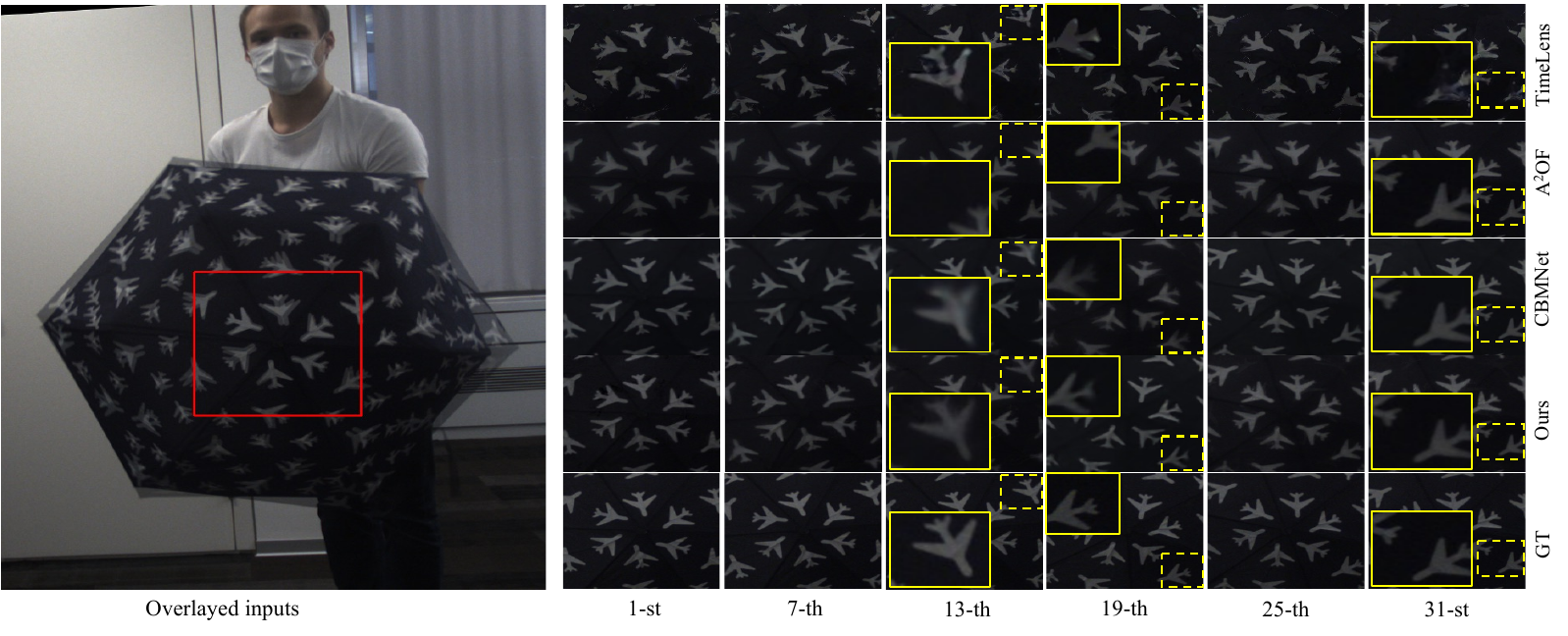}
\caption{Long-term VFI on HS-ERGB (31 frames skipped). From top to bottom: TimeLens, A$^2$OF, CBMNet, EGMR (ours) and GT. }
\label{hsergb_31}
\end{figure*}

We include a challenging stress test to our model using the HS-ERGB dataset incorporating a 31 frames skipped interpolation, as depicted in Fig. \ref{hsergb_31}. The skipping of an increased number of inter-frames intensifies the problems of motion blur and occlusion, thereby exacerbating the challenge of producing precise results. Such a large number of frame skips will greatly test the robustness of the model to complex motion. From Fig. \ref{hsergb_31}, two phenomena can be observed. First, TimeLens and CBMNet generate tearing artifacts and blurred edges, while our method maintains structural integrity and detail at these places thanks for the customized design for edge enhancement. Second, despite the apparent clarity in A$^2$OF's output, significant errors exist, such as the frames remaining static during the umbrella's rotation. The umbrella surface should complete a full rotation between two input frames, but A$^2$OF's reliance on RGB images for motion prediction can only detect minor movements when reference images have similar appearances, leading to serious interpolation errors. Conversely, our edge-guided learning strategy overcomes these challenges by leveraging both image and event modalities, providing superior adaptability across diverse application scenarios.

\subsection{Ablation Studies}
In this section, we conduct a series of experiments on the challenging dataset SNU-FILM (Extreme) to analyze the efficacy of each part contained in our proposed EGMR. 

\subsubsection{Component Efficacy} To verify the effectiveness of each component in EGMR, we set up a baseline model excluding the EGA module and the event-based visibility map ($M^e$). To construct the baseline model, we first replace EGA modules with a concatenation operation followed by convolution for OF feature fusion, and then we remove the proposed C$^2$SA module to eliminate the influence of the $M^e$.

The outcomes of the ablation studies are presented in Tab. \ref{tab: ablation}. The assessment of variants A, B, C, and D validates the efficacy of the introduced EGA module and event-based visibility map, considerably improving the baseline model's performance in VFI tasks, \eg 0.88 on PSNR and 0.011 on SSIM. Furthermore, our proposed model (EGMR), enhanced by the combination of these two designs, achieves the highest performance, demonstrating the compatibility of these two structures designed with the same edge-guided intuition. In specific, after adding the EGA module, the baseline model can better retain edge motion information from event OF, providing instructional information for motion estimation. 
Moreover, the visibility map derived from event sources are highly beneficial in addressing occlusion dilemmas. This can be ascribed to their unique property of being sensitive to moving edges, offering them a more precise depiction of occlusions that occur along moving edges, even encompassing objects absent in keyframes (as shown in Fig. \ref{fig: occlusion_mask_fig}).

\begin{table}[t] 
\centering
\caption{Impacts of the design choices of the EGMR evaluated on the SNU-FILM (Extreme) dataset.}
\renewcommand\arraystretch{1.5}
\begin{tabular}{cccccc}
\toprule[1.5pt]
\multirow{2}{*}{Variants} & \multirow{2}{*}{Baseline} & \multirow{2}{*}{C$^2$SA} & \multirow{2}{*}{EGA} & \multicolumn{2}{c}{SNU-FILM} \\ \cline{5-6} 
                          &                           &                           &                                               & PSNR          & SSIM         \\ \midrule[1pt]
  \textbf{A}        & \Checkmark              &             &                       & 29.21           & 0.901     \\
  \textbf{B}        & \Checkmark              &  \Checkmark           &                         & 29.49 & 0.906       \\
  \textbf{C}        & \Checkmark              &             &  \Checkmark                        & 29.48   &0.904       \\
  \textbf{D}        & \Checkmark              &  \Checkmark           &   \Checkmark      &            \textbf{30.09} & \textbf{0.912}  \\
\bottomrule[1.5pt]                          
\end{tabular}
\label{tab: ablation}
\end{table}

\begin{table}[t]
\centering
\caption{Attention settings of the CLA module evaluated on the SNU-FILM (Extreme) dataset.}
\renewcommand\arraystretch{1.5}
\setlength{\tabcolsep}{3pt}{
\begin{tabular}{c|cccccc}
\toprule[1.5pt]
\multicolumn{2}{c}{Attention mask} & 1$\times$1  & 3$\times$3 & 5$\times$5 & 9$\times$9 & Global \\ \midrule[1pt]
\multirow{2}{*}{SNU-FILM} & PSNR & 29.84  & \textbf{30.09}   & 29.88    & 29.82   & 29.76  \\
                     & SSIM & 0.905  & \textbf{0.912}    & 0.909    & 0.906   & 0.906 \\
\bottomrule[1.5pt]
\end{tabular}}
\label{tab: ablation_attn}
\end{table}


\subsubsection{The Impact of Attention Masks' Settings in CLA} We investigate various settings in the cross-modal local attention module (CLA) $w.r.t$ aggregating areas. We explore a series of region ($\mathcal{R}_a \in \mathbb{R}^{m \times m}$) settings of attentive aggregation from the single patch to the global area. As illustrated in Tab. \ref{tab: ablation_attn},  we can see that when the value of $m$ becomes too small ($< 3$) or large ($> 3$), the performance drops noticeably. When $m$ is small, the CLA has difficulty optimizing frame OF using correlated event OF because the aggregation range is too narrow. Conversely, when $m$ is too large, the CLA tends to accept all biases from the event OF, including uncorrelated and noisy OF features, resulting in worse predictions. In this work, we experimentally set $m$ as 3.

\section{Conclusion}
This paper proposes an end-to-end E-VFI framework (EGMR) for event-based video frame interpolation via edge guided motion refinement. Given that event data primarily supplies high-confidence features at scene edges, we incorporates an Edge Guided Attentive (EGA) module to rectify estimated video motion via attentive aggregation based on local correlation of cross-modal features in a coarse to fine strategy. Furthermore, we introduce an event-based visibility map to effectively address the occlusion problem in VFI tasks by complement precise inter-frame visual references in the warping refinement process. Comprehensive experiments on both synthetic and real datasets validate the advantages of our proposed EGMR.

\bibliographystyle{IEEEtran}
\bibliography{IEEEabrv,Bibliography}

\vfill

\end{document}